\documentclass{files/ecai} 

\usepackage{latexsym}
\usepackage{amssymb}
\usepackage{amsmath}
\usepackage{amsthm}
\usepackage{booktabs}
\usepackage{enumitem}
\usepackage{graphicx}
\usepackage{color}
\usepackage[hidelinks]{hyperref}
\usepackage{algorithm}
\usepackage{algpseudocode}
\usepackage{subfig} 
\usepackage[most]{tcolorbox}
\usepackage{booktabs}
\usepackage{makecell}

\newtcolorbox{summary}[1][]{colback=orange!10!white, colframe=orange!80!black, title=#1}

\newcommand{\BibTeX}{B\kern-.05em{\sc i\kern-.025em b}\kern-.08em\TeX}

\begin{document}


\begin{frontmatter}


\paperid{123} 


\title{
Human-LLM Synergy in Context-Aware Adaptive Architecture 
for Scalable Drone Swarm Operation
}

%

%
\author[A]{\fnms{Ahmed R.}~\snm{Sadik}\orcid{0000-0001-8291-2211}\thanks{Corresponding Author. Email: ahmed.sadik@honda-ri.de.}}
\author[B]{\fnms{Muhammad}~\snm{Ashfaq}\orcid{0000-0003-1870-7680}}
\author[B]{\fnms{Niko}~\snm{M\"{a}kitalo}\orcid{0000-0002-7994-3700}}
\author[B]{\fnms{Tommi}~\snm{Mikkonen}\orcid{0000-0002-8540-9918}}

\address[A]{Honda Research Institute Europe, Germany}
\address[B]{University of Jyv\"{a}skyl\"{a}, Finland}

\begin{abstract}

The deployment of autonomous drone swarms in disaster response missions necessitates the development of flexible, scalable, and robust coordination systems. 
Traditional fixed architectures struggle to cope with dynamic and unpredictable environments, leading to inefficiencies in energy consumption and connectivity. 
This paper addresses this gap by proposing an adaptive architecture for drone swarms, leveraging a Large Language Model~(LLM) to dynamically select the optimal architecture---centralized, hierarchical, or holonic---based on real-time mission parameters such as task complexity, swarm size, and communication stability. 
Our system addresses the challenges of scalability, adaptability, and robustness, ensuring efficient energy consumption and maintaining connectivity under varying conditions.
Extensive simulations demonstrate that our adaptive architecture outperforms traditional static models in terms of scalability, energy efficiency, and connectivity. These results highlight the potential of our approach to provide a scalable, adaptable, and resilient solution for real-world disaster response scenarios.

\end{abstract}

\end{frontmatter}

\begin{figure*}[!ht]
    \centering
    \includegraphics[width=0.8\textwidth]{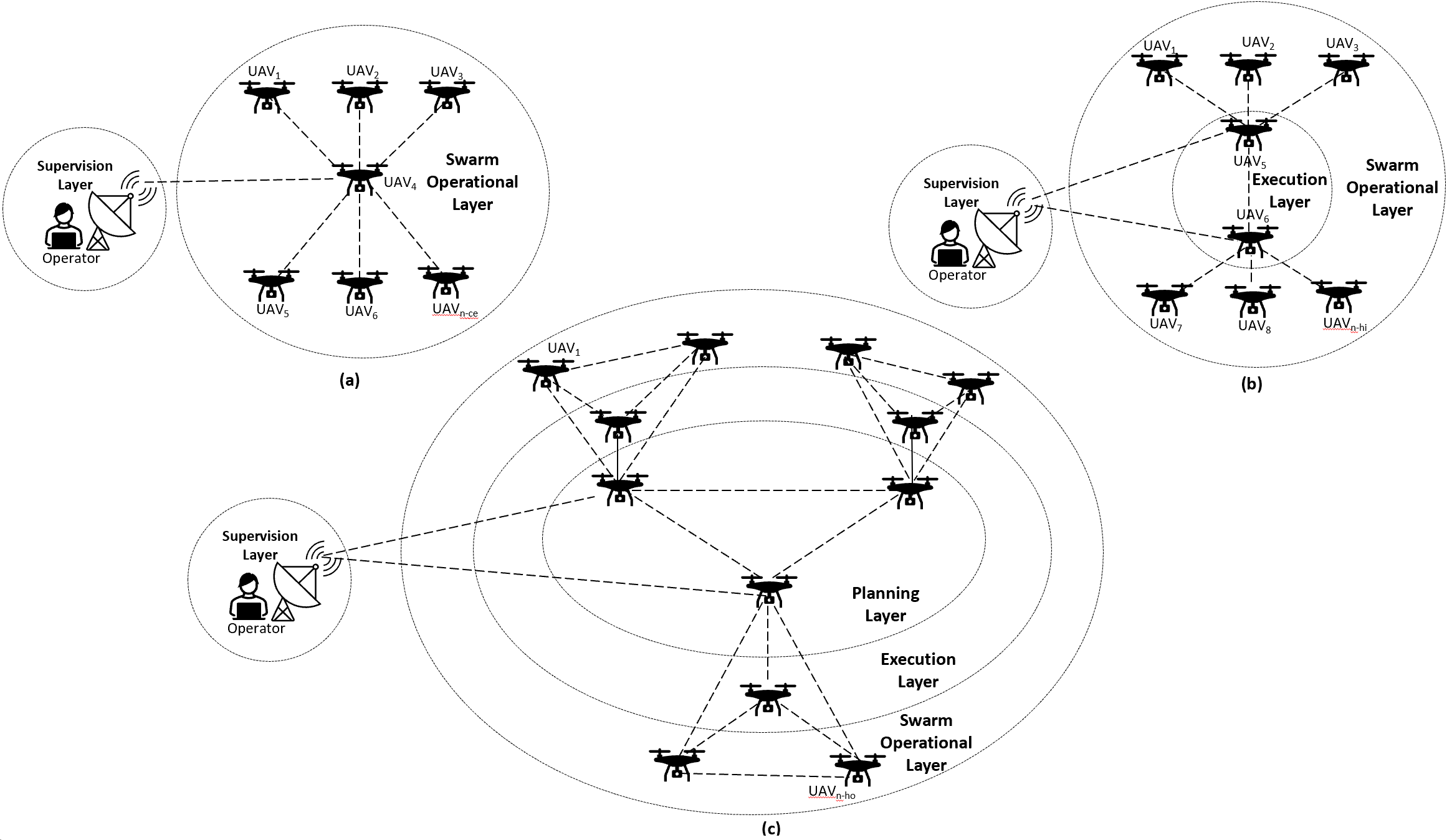}
    \caption{Three primary drone swarm architectures: (a) Centralized, (b) Hierarchical, and (c) Holonic.} \vspace{2em}
    \label{fig:architectures}
\end{figure*}

\section{Introduction}

Autonomous drone swarms are increasingly deployed across diverse applications, including surveillance ~\citep{nedjah2006swarm}, environmental monitoring~\citep{shi2012innovations}, disaster response~\citep{tan2020advances}, and logistics~\citep{del2019swarm}. Among these, disaster response presents unique operational challenges ~\citep{alexander2007disaster}, as unpredictable environmental conditions ~\citep{zedadra2018towards}, damaged communication infrastructure ~\citep{hasan2022swarm}, and fluctuating mission priorities demand a high degree of adaptability and resilience~\citep{Bakirci2023Surveillance}. 
Traditional drone swarm coordination approaches rely on fixed architectures that struggle to adapt efficiently to dynamic scenarios~\citep{Chen2020Toward}.

As illustrated in Fig.~\ref{fig:architectures}, drone swarms can operate under three primary architectures: \textit{centralized}, \textit{hierarchical}, and \textit{holonic}~\citep{sadik2023self}. 
Each architecture has distinct benefits and limitations, making the selection process highly dependent on mission requirements and environmental constraints. The centralized architecture follows a one-layer pattern, where all drones communicate directly with a \hypertarget{MCC}{mission control center (MCC)} supervised by a human operator. While this structure enables low communication delay and minimal data traffic, it suffers from a single point of failure, limited scalability, and constrained collaboration among drones~\citep{Hu2018To}.

The hierarchical architecture introduces an execution layer, where lead drones act as intermediaries between the \hyperlink{MCC}{\normalcolor MCC} and the swarm.
This structure improves scalability, extends communication range, and enhances cooperation~\citep{Xu2021Communication}. However, it introduces multiple points of failure, increases network traffic, and results in longer communication delays~\citep{Tahir2020Comparison}. Inspired by Industry 4.0 and smart manufacturing, the holonic architecture~\citep{sadik2016novel}, represents the most advanced swarm coordination model. It employs a multi-layer hierarchy, allowing drones to dynamically form teams based on peer-to-peer collaboration. This design offers high resilience, enhanced scalability, and self-organizing capabilities, making it suitable for large-scale missions~\citep{Sacco2023Towards, Jiang2022AIAssisted}. However, its complexity introduces longer delays and increased communication traffic~\citep{Jiang2022AIAssisted, Muller2022Architecture}.

Each swarm architecture is suited to specific scenarios: centralized control is effective for initial deployment and task assignment, hierarchical control facilitates coordinated \hypertarget{SAR}{Search-And-Rescue (SAR)} operations, and holonic organization ensures resilience in communication-degraded environments. However, no single architecture is universally suitable, as disaster response missions involve diverse and unpredictable conditions that demand varying approaches~\citep{botterill2005disaster}. A static approach fails to accommodate the shifting demands of real-world missions, particularly in unpredictable environments. The research gap, therefore, lies in enabling dynamic transitions between architectures, allowing systems to adapt seamlessly to changing conditions and maintain operational effectiveness. To address this challenge, we propose a \hypertarget{LLM}{Large Language Model (LLM)}-driven decision-making system that assists the operator in selecting the optimal architecture, monitoring swarm conditions, and suggesting real-time architecture adjustments.
The system enables bidirectional interaction, where the operator provides mission objectives, and the \hyperlink{LLM}{\normalcolor LLM} determines the most effective control paradigm based on environmental conditions, energy constraints, and communication stability. This adaptive approach ensures that drone swarms operate with greater efficiency, resilience, and scalability while reducing operator workload~\citep{brulin2025system}.

In this article, we first introduce the adaptive architecture and its dynamic communication structure selection mechanism. Next, we present the swarm model and its key parameters. We then walk through the simulation setup and results, including performance metrics and visualizations. Finally, we conclude with key findings and future research directions.

\section{LLM-Driven adaptive architecture concept} 

\label{sec:proposed-architecture}

\begin{figure}[!ht]
    \centering
    \includegraphics[width=\linewidth]{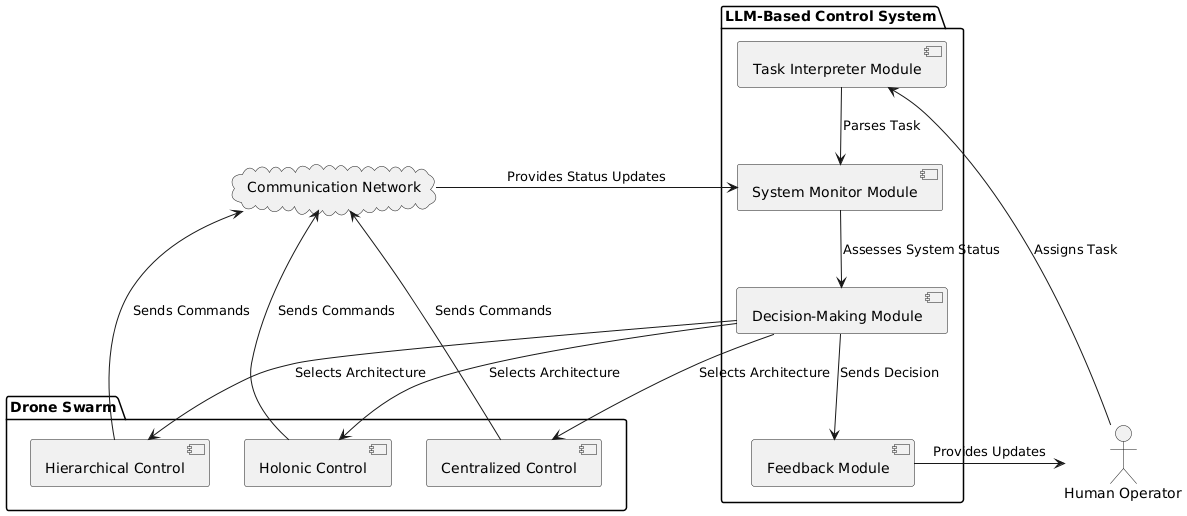}
    \caption{LLM-Driven adaptive architecture concept} \vspace{2em}
    \label{fig:system_architecture}
\end{figure}

\vspace{1em} 

Our proposed \hyperlink{LLM}{\normalcolor LLM-driven} adaptive system monitors mission conditions, swarm dynamics, and environmental factors and selects the most suitable control paradigm (centralized, hierarchical, and holonic), ensuring optimal performance in unpredictable disaster response scenarios.

As illustrated in Fig.~\ref{fig:system_architecture}, the proposed system consists of four key components. The Human Operator serves as the primary decision-maker, assigning high-level mission objectives and receiving real-time feedback on the swarm's status. 
The LLM-Based Control System interprets tasks, monitors the system, selects control architectures, and provides recommendations to the operator.
The Drone Swarm operates under one of three control architectures---centralized, hierarchical, or holonic---depending on the task, system constraints, and environmental conditions.
Finally, the Communication Network facilitates bidirectional data exchange between the LLM-based control system and the drone swarm. 

\begin{table*}
\vspace*{4pt}
\caption{LLM-Driven Architecture Recommendations for Task-Based Selection and System Monitoring}
\label{tbl:recommendations}
\small
\begin{tabular*}{\textwidth}{@{}p{170pt}p{40pt}p{70pt}p{50pt}p{70pt}@{}}
\toprule
\textbf{Scenario/Status} & \textbf{Swarm Size} & \textbf{Communication Quality} & \textbf{Failure Probability} & \textbf{Recommended Architecture} \\
\midrule
\multicolumn{5}{c}{\textbf{Task-Based Recommendations}} \\
\midrule
Search and Rescue & Small & Good & Low & Centralized \\
Search and Rescue & Large & Low & High & Holonic \\
Large-Area Mapping & Small & Good & Low & Hierarchical \\
Large-Area Mapping & Large & Moderate & Moderate & Holonic \\
Emergency Supply Delivery & Medium & Good & Low & Hierarchical \\
Emergency Supply Delivery & Large & Low & High & Holonic \\
Post-Disaster Environmental Assessment & Medium & Good & Low & Hierarchical \\
Post-Disaster Environmental Assessment & Large & Low & High & Holonic \\
\midrule
\multicolumn{5}{c}{\textbf{Status-Based Recommendations}} \\
\midrule
Critical failure & Small & Low & High & Hierarchical \\
Idle state & Small & Good & Low & Centralized \\
Spread out & Medium & Moderate & Moderate & Hierarchical \\
Overload & Large & Low & High & Holonic \\
\bottomrule
\end{tabular*}
\vspace*{8pt}
\end{table*}

\subsection{Workflow for Dynamic Architecture Selection}

The proposed system follows a structured, adaptive decision-making process to ensure optimal architecture selection. The process is divided into six key phases, described below.The process begins with the Human Operator assigning a task to the LLM-Based Control System. These tasks may include aerial surveillance, victim localization, or emergency supply delivery, depending on mission requirements. Once the task is assigned, it is passed to the system for processing.

The Task Interpreter Module processes the mission objective and forwards it to the System Monitor Module, which continuously evaluates key system parameters. These parameters include drone energy levels to determine the feasibility of long-duration missions; communication stability, which assesses connectivity for centralized or hierarchical control; swarm size, which evaluates whether the number of drones aligns with the mission requirements and control strategy; and processing load, which determines the computational demands required for real-time decision-making \citep{sendhoff2020cooperative}.Based on the parameters evaluated during system monitoring, the Decision-Making Module selects the most appropriate control architecture. Centralized control is applied when strong communication is available, typically for small swarm sizes with high operator oversight. Hierarchical control is used when scalability is necessary, distributing responsibilities across different levels of the swarm. Holonic control is activated in environments where drones must self-organize due to unpredictable conditions or communication limitations ~\citep{brulin2025system}.

After the appropriate architecture is selected, the control module sends operational commands to the drone swarm via the Communication Network. This initiates the task execution phase, during which the drones perform the assigned mission collaboratively and autonomously, depending on the selected architecture. Throughout task execution, the Drone Swarm provides continuous status updates through the Communication Network back to the System Monitor Module. This real-time feedback loop allows for ongoing monitoring of the swarm's state and mission progress, enabling proactive or corrective interventions if required \citep{sendhoff2020cooperative}.

Finally, if environmental or operational conditions change—such as a drop in battery levels, increased failure probability, or degraded communication quality—the LLM dynamically reconfigures the swarm's control architecture. This reconfiguration is communicated to the Human Operator, ensuring transparency and continuity of mission control. This real-time adaptation guarantees resilience and robust performance in complex disaster response scenarios, a critical feature in modern AI-driven autonomous systems.

\subsection{Integrated Human-LLM Decision Making}

The proposed LLM-driven control system facilitates bidirectional interaction between the human operator and the LLM decision-making module, ensuring optimal drone swarm management in disaster response operations \citep{sendhoff2020cooperative}.

On one hand, the operator assigns a task, and the LLM selects the most suitable architecture based on constraints such as swarm size, communication quality, and drone failure probability. Table~\ref{tbl:recommendations} illustrates this decision-making process, showing how the LLM evaluates these constraints to determine the optimal architecture for each task. For example, the system selects a Centralized architecture for small swarms with good communication to optimize coordination. For larger swarms or those with moderate communication, it switches to Hierarchical or Holonic architectures to maintain scalability and energy efficiency, as demonstrated in the table.

On the other hand, the LLM continuously monitors swarm conditions and dynamically adjusts the architecture to optimize performance, scalability, and resilience. It selects between Centralized, Hierarchical, or Holonic architectures in real-time, based on evolving operational constraints. As shown in Table~\ref{tbl:recommendations}, the LLM favors Centralized control for small swarms and employs Hierarchical or Holonic architectures as swarm size increases. Communication quality also guides architecture choice: Centralized and Hierarchical architectures are preferred in conditions with good or moderate communication, while Holonic is favored when communication is poor or unstable. Furthermore, Holonic architecture is ideal in high failure scenarios due to its decentralized and fault-tolerant nature, whereas Hierarchical provides a balance of coordination and robustness under moderate failure probabilities

\section{Algorithmic Foundations of Swarm Operation}

Disaster response operations demand rapid deployment of \hyperlink{SAR}{\normalcolor SAR} missions, environmental assessments, emergency communications, and supply deliveries. Drone swarms offer a scalable and adaptable solution to these challenges. However, their effectiveness depends on the ability to dynamically adapt to environmental conditions, system constraints, and evolving mission priorities. 

This study investigates an LLM-based decision-making system that enables a drone swarm to transition seamlessly between centralized, hierarchical, and holonic control architectures in response to real-time operational demands. The drones perform diverse tasks, including SAR for survivor localization, large-scale structural risk mapping, emergency supply delivery, post-disaster environmental monitoring, and the establishment of communication relays in disrupted areas. The simulation evaluates scalability, energy consumption, and connectivity by introducing two additional drones in each iteration, while accounting for system constraints. Each drone is equipped with a fixed battery capacity of 700\,W. Energy consumption is divided into fixed (flight and task execution) and variable (communication) components.

In the centralized architecture, all drones communicate with a central node, leading to significant communication overhead. Energy consumption due to communication scales linearly with swarm size, as in the worst-case scenario a drone must wait for all others to transmit (assuming a first-come, first-served scheduling mechanism). Under these assumptions, the fixed operational energy is \(K_o = 10\,\text{W}\), and communication energy is given by:
\begin{equation*}
E_{\text{centralized}} = K_o + K_{c\text{-}ce} \cdot N, \quad \text{where } K_{c\text{-}ce} = 5\,\text{W}.
\end{equation*}

In the hierarchical architecture, drones are organized into clusters, with cluster heads managing intra- and inter-cluster communication~\citep{lindsey2002pegasis}. This reduces communication overhead, with energy consumption growing sub-linearly, assuming the number of clusters scales with the square root of the swarm size:
\begin{equation*}
E_{\text{hierarchical}} = K_o + K_{c\text{-}hi} \cdot \sqrt{N}, \quad \text{where } K_{c\text{-}hi} = 3\,\text{W}.
\end{equation*}

In the holonic architecture, drones communicate only with their immediate neighbors~\citep{brambilla2013swarm}, resulting in constant communication energy regardless of swarm size:
\begin{equation*}
E_{\text{holonic}} = K_o + K_{c\text{-}ho}, \quad \text{where } K_{c\text{-}ho} = 1\,\text{W}.
\end{equation*}

To analyze the scalability of each control architecture, the simulation enforces one of the three predefined modes (centralized, hierarchical, or holonic) and evaluates swarm performance under constrained conditions. In hierarchical and holonic architectures, structured communication becomes active only after reaching a minimum swarm size—14 drones for hierarchical and 42 for holonic. The experiment monitors swarm size based on battery depletion and replaces any drone that runs out of battery in each iteration.

Algorithm~\ref{alg:swarm-operation} simulates the operation of the swarm under adaptive architecture compared with the three static architectures.

\begin{figure*}[!htbp]
    \centering
    \subfloat[Scalability]{
        \label{fig:Scalability}
        \includegraphics[width=0.4\textwidth]{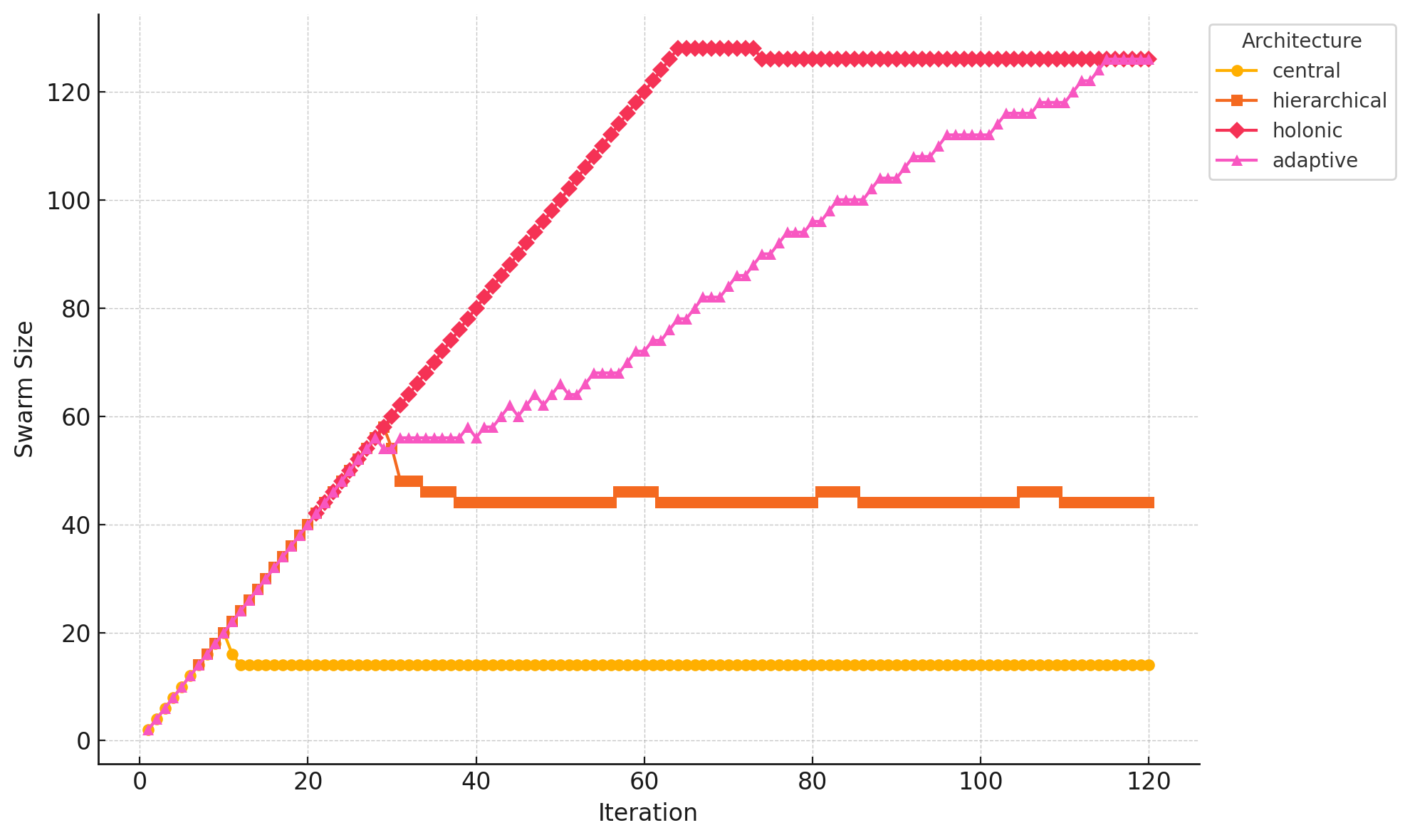}
    }%
    ~
    \subfloat[Connectivity]{
        \label{fig:Connectivity}
        \includegraphics[width=0.4\textwidth]{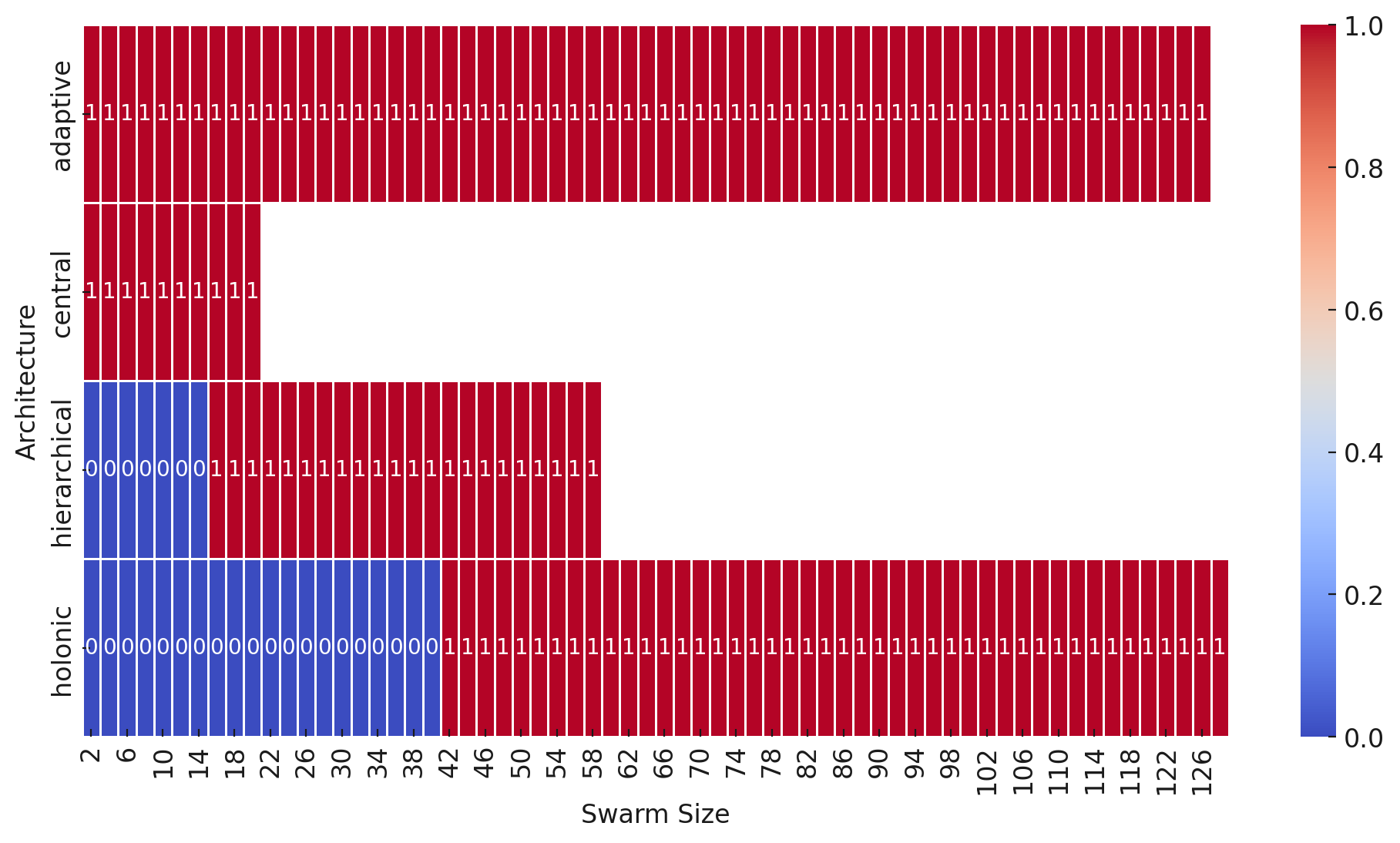}
    }
    \caption{Scalability vs Connectivity across architectures} \vspace{0.8em}
    \label{fig:scalability-vs-connectivity}
\end{figure*}

\begin{figure*}[t!]
    \centering
    \subfloat[Energy consumption over iterations]{
        \label{fig:EnergyConsumpationPerIteration}
        \includegraphics[width=0.4\textwidth]{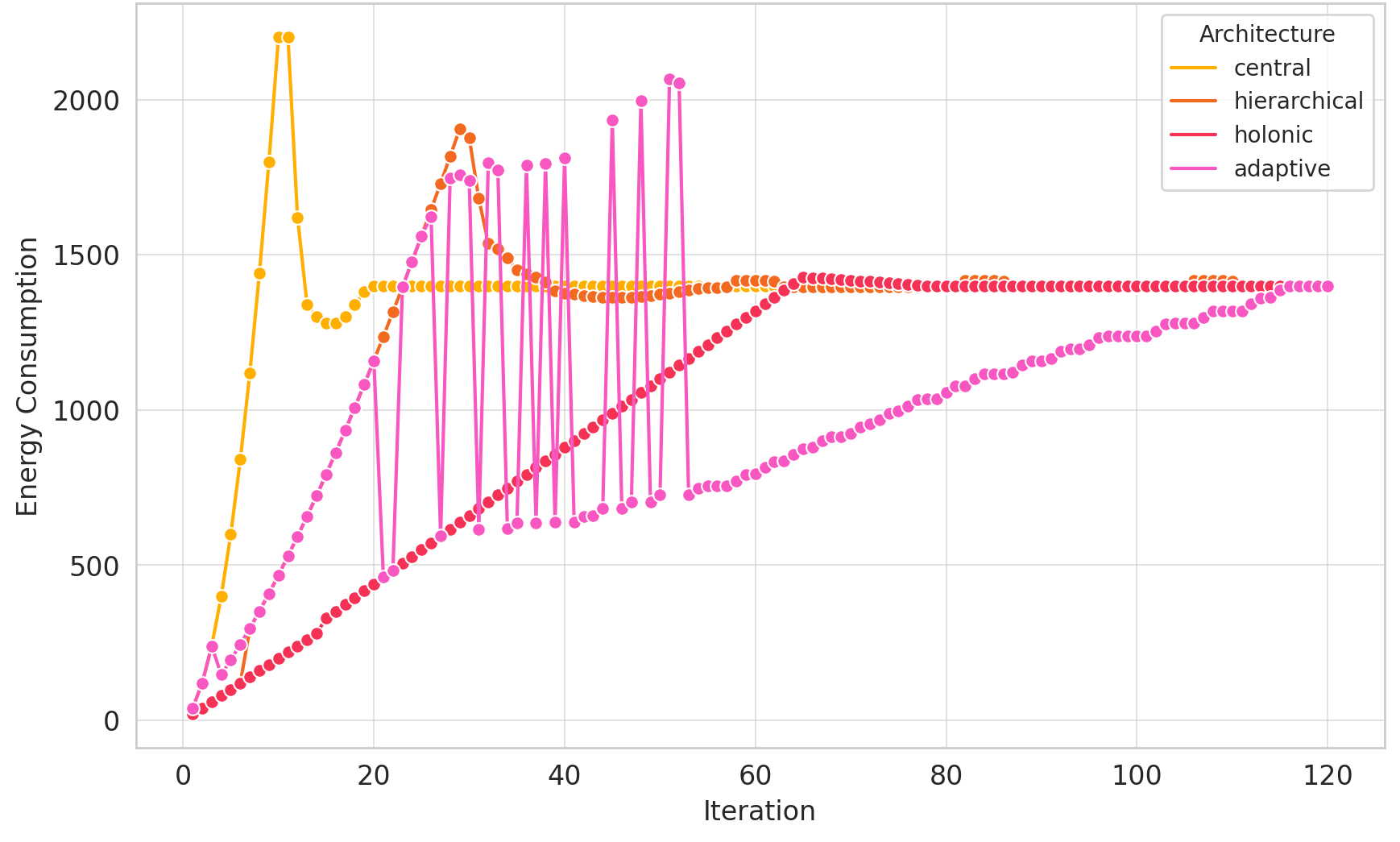}
    }%
    ~
    \subfloat[Energy consumption distribution]{
        \label{fig:EnergyConsumptionDistribution}
        \includegraphics[width=0.4\textwidth]{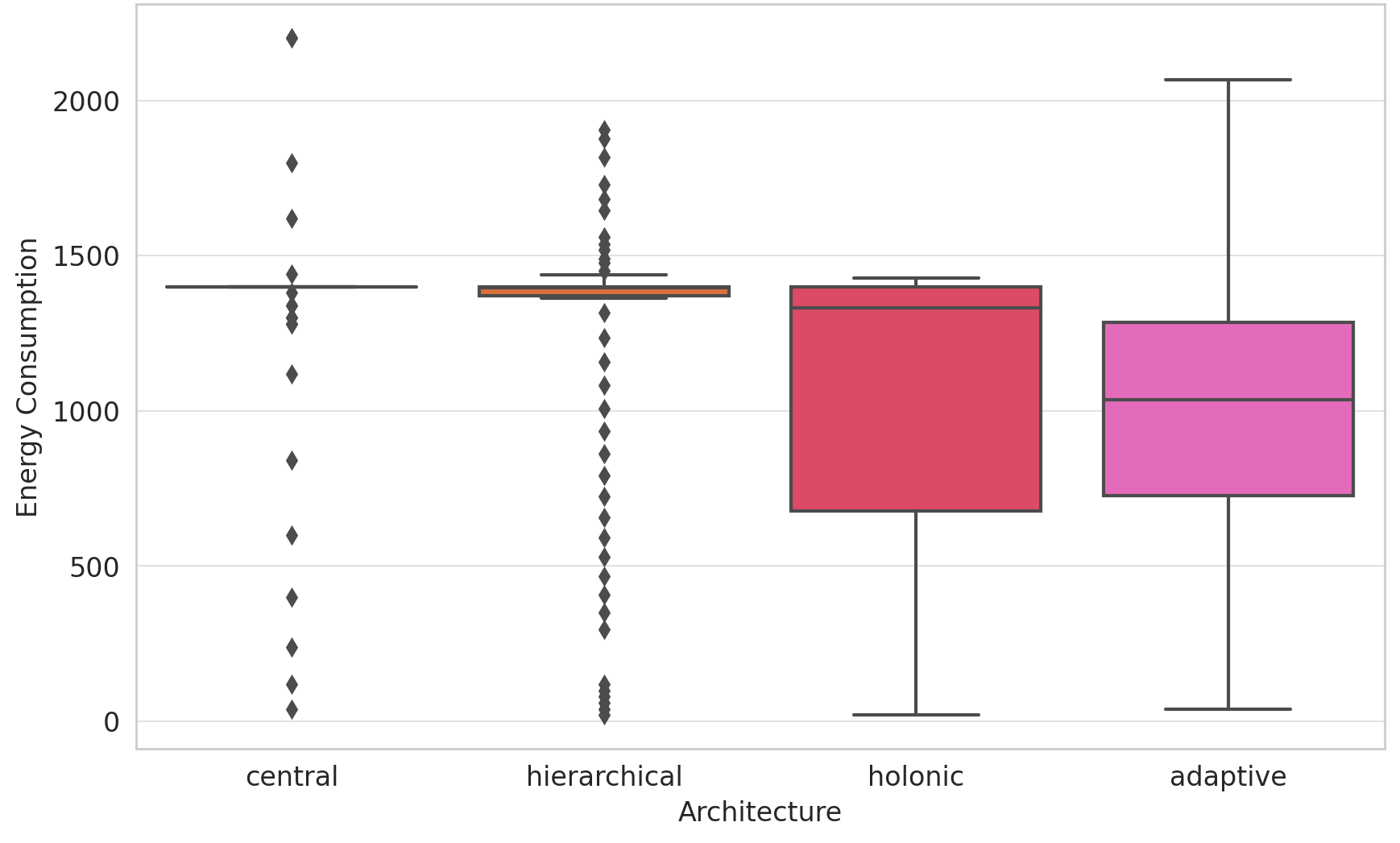}
    }
    \caption{Comparison of Energy Consumption Across Architectures} \vspace{2em}
    \label{fig:EnergyConsumption}
\end{figure*}

\begin{algorithm}[H]
    \caption{Swarm Operation with Battery Tracking}
    \label{alg:swarm-operation}
    \begin{algorithmic}[1]
        \Require Initial swarm size $N_0$, battery capacity $B = 700W$, energy consumption model $E_{arch}$, minimum formation sizes: $N_{hierarchical} = 14$, $N_{holonic} = 42$
        \State Initialize swarm with $N_0$ drones, each with $B = 700W$
        \For {each iteration}
            \State Add two new drones with $B = 700W$: $N \gets N + 2$
            \For {each drone $d_i$ in swarm}
                \State Determine control architecture:
                \If {Selected Control = Centralized}
                    \State Assign \textbf{Centralized Control}: $E_i = 10\,W + 5\,W \cdot N$
                \ElsIf {Selected Control = Hierarchical}
                    \If {$N \geq N_{hierarchical}$}
                        \State Form \textbf{Hierarchical Control}
                        \State $E_i = 10\,W + 3\,W \cdot \sqrt{N}$
                    \Else
                        \State Drones cannot communicate
                        \State $E_i = 10\,W$
                    \EndIf
                \ElsIf {Selected Control = Holonic}
                    \If {$N \geq N_{holonic}$}
                        \State Form \textbf{Holonic Control}
                        \State $E_i = 10\,W + 1\,W$
                    \Else
                        \State Drones cannot communicate
                        \State $E_i = 10\,W$
                    \EndIf
                \EndIf
                \State Update battery: $B_i \gets B_i - E_i$
            \EndFor
            \State Track depleted drones: $D_{\text{depleted}} = \{d_i \mid B_i \leq 0\}$
            \State Remove $|D_{\text{depleted}}|$ drones from swarm: $N \gets N - |D_{\text{depleted}}|$
        \EndFor
    \end{algorithmic}
\end{algorithm}

\section{Performance Evaluation}

This section presents the simulation-based performance analysis of the four swarm control architectures: centralized, hierarchical, holonic, and adaptive. The evaluation focuses on two critical dimensions: (i) scalability and connectivity, which assess how effectively each architecture supports swarm growth and maintains communication integrity; and (ii) energy efficiency, which measures how well each architecture manages power consumption as the swarm expands over time. The results are derived from controlled simulation experiments and are presented through comparative visualizations and summary statistics to highlight architectural trade-offs and operational implications.

\subsection{Scalability vs Connectivity}

Scalability and connectivity are two fundamental performance metrics in swarm operations. Scalability refers to a swarm’s ability to grow in size while maintaining operational efficiency, whereas connectivity indicates the extent to which all agents remain in effective communication. Figure~\ref{fig:scalability-vs-connectivity} provides a visual comparison of these attributes across four swarm architectures, and Table~\ref{tab:scalability_connectivity} summarizes key simulation metrics, including the iteration at which connectivity is established, the maximum swarm size (growth limit), and the saturation point.

\begin{table}[h]
    \centering
    \begin{tabular}{lccc}
        \toprule
        \textbf{Architecture} & \makecell{\textbf{Connected} \\ \textbf{From (D@i)}} & \makecell{\textbf{Growth} \\ \textbf{Limit (D)}} & \makecell{\textbf{Saturation} \\ \textbf{(D@i)}} \\
        \midrule
        Adaptive      & 2@0    & 126  & 126@115  \\
        Centralized   & 2@0    & 20   & 14@12   \\
        Hierarchical  & 14@7   & 58   & 44@38   \\
        Holonic       & 42@21  & 128  & 126@74  \\
        \bottomrule
    \end{tabular}
    \caption{Scalability and Connectivity Performance Across Architectures}
    \label{tab:scalability_connectivity}
\end{table}

The Adaptive architecture demonstrates superior scalability and robustness by starting with 2 drones at iteration~0 and growing steadily to 126 drones, reaching saturation at iteration~115. It maintains full connectivity throughout, highlighting exceptional flexibility and fault tolerance. In contrast, the Holonic architecture only establishes connectivity at 42 drones (iteration~21) but scales rapidly to a maximum of 128 drones, stabilizing at 126 drones around iteration~74; however, its localized communication model may lead to connectivity challenges in smaller configurations. The Hierarchical architecture achieves connectivity at 14 drones (iteration~7) and scales moderately to 58 drones before plateauing at 44 drones by iteration~38, reflecting a balance between scalability and connectivity that is ultimately limited by communication overhead. Finally, the Centralized architecture, although connected from the start (2@0), exhibits constrained growth—peaking at only 20 drones and saturating at 14 drones by iteration~12—which makes it unsuitable for large-scale swarm deployments.

In summary, the Adaptive architecture clearly outperforms the others in both scalability and connectivity, while the Holonic model offers strong scaling after an initial threshold, the Hierarchical approach provides a balanced but limited performance, and the Centralized system, despite its consistent connectivity, lacks scalability.

\subsection{Energy Efficiency Evaluation: Progress vs Distribution}

Energy efficiency measures how effectively each swarm architecture manages energy consumption as the swarm grows and operates over time. This evaluation considers two perspectives: (a) the evolution of energy consumption over iterations and (b) the statistical distribution of energy usage across all iterations and swarm sizes. Figure~\ref{fig:EnergyConsumption}—with subfigure (a) depicting energy consumption over iterations and subfigure (b) showing the distribution via box plots—visually compares these aspects, while Table~\ref{tab:energy_consumption} provides key statistics such as median energy consumption, variance, and peak energy usage.

\begin{table}[h]
    \centering
    \begin{tabular}{lccc}
        \toprule
        \textbf{Architecture} & \makecell{\textbf{Median Energy} \\ \textbf{(W)}} & \textbf{Variance} & \makecell{\textbf{Peak Energy} \\ \textbf{(W)}} \\
        \midrule
        Adaptive      & 1036.0  & 185{,}300.95  & 2065.0  \\
        Centralized   & 1400.0  & 69{,}102.41   & 2200.0  \\
        Hierarchical  & 1396.0  & 146{,}639.89  & 1905.0  \\
        Holonic       & 1331.0  & 210{,}803.67  & 1428.0  \\
        \bottomrule
    \end{tabular}
    \caption{Energy Consumption Statistics Across Architectures}
    \label{tab:energy_consumption}
\end{table}

Figure~\ref{fig:EnergyConsumpationPerIteration} shows that the Centralized architecture exhibits an early sharp spike in energy usage, peaking at 2200~W and then stabilizing, which underscores its high overhead and limited scalability. The Hierarchical model increases more gradually, peaking at 1905~W with smoother behavior over time. The Holonic architecture displays steady growth with a relatively low peak of 1428~W, suggesting effective energy management despite the highest observed variance. In contrast, the Adaptive architecture, although peaking at 2065~W, maintains the lowest median energy consumption at 1036~W and exhibits lower variance compared to the Holonic model, thereby proving to be the most energy-efficient and consistent across varying swarm sizes.

In conclusion, the Adaptive architecture is best suited for large-scale swarm operations owing to its superior energy efficiency and consistency. The Holonic model, while energy-efficient at scale, experiences significant fluctuations; the Hierarchical approach offers a trade-off between energy stability and consumption; and the Centralized architecture, though predictable, is both energy-intensive and less scalable.

\subsection{Overall Performance Comparison}

\begin{figure}[!ht]
    \centering
    \includegraphics[width=0.8\linewidth]{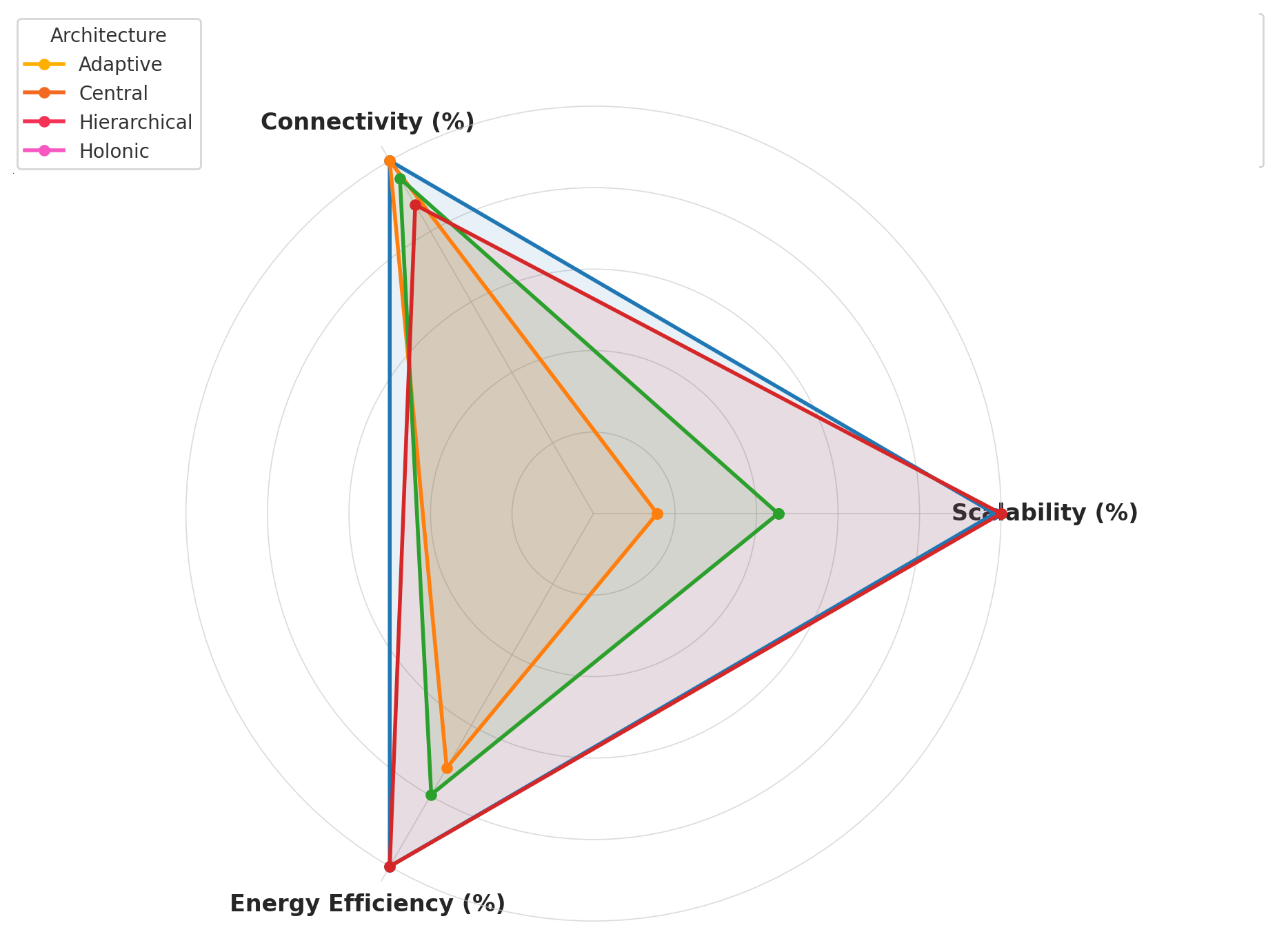}
    \caption{Performance Matrices Comparison} \vspace{2em}
    \label{fig:overall_performance}
\end{figure}

To provide a holistic assessment, the overall performance of the four architectures—adaptive, holonic, hierarchical, and centralized—is compared across three key dimensions: scalability, connectivity, and energy efficiency. Figure~\ref{fig:overall_performance} presents a radar chart summarizing the relative strengths and weaknesses of each architecture in these categories. The scores are derived from normalized metrics based on the simulation data presented in the previous subsections.

The Adaptive architecture emerges as the most balanced and high-performing solution. It achieves the highest scores across all three metrics, maintaining full connectivity across all swarm sizes, scaling efficiently to large swarm populations, and demonstrating the lowest median energy consumption. Its superior energy efficiency and consistent performance make it particularly well-suited for large-scale, long-duration swarm operations.

The Holonic architecture performs competitively in terms of scalability and energy efficiency, reaching the highest swarm sizes with relatively low energy peaks. However, it lags in connectivity, especially during early swarm formation, due to its reliance on localized communication, which becomes less effective as the network topology becomes sparse.

The Hierarchical architecture provides a moderate balance across all dimensions. It scales better than centralized systems and maintains a relatively stable energy profile, but its connectivity weakens beyond mid-sized swarms. While it offers a reasonable trade-off between performance metrics, it ultimately plateaus earlier than the adaptive and holonic models.

The Centralized architecture demonstrates strong connectivity, maintaining full network cohesion across all swarm sizes. However, it is the least scalable due to early saturation and sharp increases in energy consumption. Its rigid communication model incurs high overhead, making it unsuitable for dynamic or large-scale swarm environments.

Figure~\ref{fig:overall_performance} reinforces the findings from earlier analyses. The Adaptive architecture consistently outperforms others by balancing scalability, connectivity, and energy efficiency. Holonic follows closely in scalability and efficiency but suffers from connectivity limitations. Hierarchical achieves moderate performance across the board, while Centralized architecture, despite its consistent connectivity, underperforms in scalability and energy management.

\section{Implications and Future Directions}

This paper presents an LLM-driven adaptive architecture for autonomous drone swarms, demonstrating significant advantages in scalability, energy efficiency, and resilience over traditional static models. By dynamically selecting the most suitable communication paradigm based on real-time mission parameters—such as swarm size, communication quality, and drone failure probability—the adaptive architecture ensures that drones maintain optimal performance even under fluctuating environmental and operational conditions.

The proposed approach offers multiple key benefits. First, it facilitates human-AI synergy by allowing the operator to maintain high-level oversight while delegating adaptive decision-making to the LLM, ensuring that the system responds to changing conditions in real time. Second, the architecture enhances energy efficiency by matching control strategies to mission needs, avoiding unnecessary processing overhead. Third, it ensures resilience to drone failures through timely transitions to fault-tolerant modes such as holonic control. Fourth, it enables scalability across various swarm sizes by adapting control complexity accordingly. Finally, the LLM-based architecture increases mission reliability, reducing delays and improving safety in time-sensitive disaster response scenarios.

Simulation results validate the advantages of adaptability. The centralized model, while simple, suffers from high communication overhead, limiting its scalability. The hierarchical model offers better distribution of control but remains suboptimal in energy use. The holonic model is efficient but suffers from reduced global awareness in large swarms. The adaptive approach, however, consistently outperforms these static models by dynamically selecting the most appropriate architecture in context, thereby balancing communication efficiency, energy usage, and coordination.

Beyond its immediate benefits, this work contributes to the broader challenge of designing flexible distributed systems for the internet era. AI-augmented flexibility enables autonomous systems to dynamically reconfigure in response to changing conditions, a principle increasingly important for distributed cyber-physical systems. Recent studies support integrating LLMs into swarm intelligence as a means to enhance scalability, resilience, and efficiency in multi-agent environments. For instance, AI-driven reconfiguration improves decision accuracy and reaction times, while self-organizing communication structures enable adaptive routing and resource optimization.

To bring this approach closer to real-world deployment, future research should explore the integration of additional environmental parameters into the decision-making process, such as terrain topology, weather conditions, and dynamic mission objectives. Incorporating online learning mechanisms could enable the LLM to adapt its architecture selection strategies based on accumulated experience. Hybrid architectures that blend centralized coordination with decentralized autonomy may also offer improved performance in heterogeneous mission profiles. Moreover, hardware-in-the-loop simulations and real-world field trials will be essential for validating the system's robustness, particularly in high-stakes scenarios such as disaster relief, emergency communication relay, and environmental monitoring.

By implementing our proposed LLM-driven adaptive architecture selection, drone swarms gain self-organizing capabilities that empower them to autonomously manage complex, dynamic missions with minimal human intervention. This framework represents a significant step toward intelligent, responsive, and scalable multi-agent systems for next-generation autonomous operations.



\bibliography{main}

\end{document}